\documentclass{article}

\usepackage[preprint, nonatbib]{neurips_2024}

\usepackage[utf8]{inputenc} 
\usepackage[T1]{fontenc}    
\usepackage{hyperref}       
\usepackage{url}            
\usepackage{booktabs}       
\usepackage{amsfonts}       
\usepackage{nicefrac}       
\usepackage{microtype}      
\usepackage{xcolor}         

\usepackage{graphicx}
\usepackage{algorithm} 
\usepackage{algpseudocode} 
\usepackage{amsmath}
\usepackage{xcolor}
\usepackage{setspace}
\usepackage{lineno}
\usepackage{subcaption}

\title{Resource Governance in Networked Systems via Integrated Variational Autoencoders and Reinforcement Learning}

\author{Qiliang Chen \\MAGICS lab \\ College of Engineering \\
Northeastern University \\
  Boston, MA 02115, USA \\
  \textit{chen.qil@northeastern.edu} \\\And
  Babak Heydari*\\MAGICS lab \\ 
  College of Engineering and Network Science Institute \\
  Northeastern University \\
  Boston, MA 02115, USA \\
  \textit{b.heydari@northeastern.edu} \\}

\begin{document}

\maketitle

\begin{abstract}
We introduce a framework that integrates variational autoencoders (VAE) with reinforcement learning (RL) to balance system performance and resource usage in multi-agent systems by dynamically adjusting network structures over time. A key innovation of this method is its capability to handle the vast action space of the network structure. This is achieved by combining Variational Auto-Encoder and Deep Reinforcement Learning to control the latent space encoded from the network structures. The proposed method, evaluated on the modified OpenAI particle environment under various scenarios, not only demonstrates superior performance compared to baselines but also reveals interesting strategies and insights through the learned behaviors.
\end{abstract}


\onehalfspacing
\section{Introduction}

Modern complex systems increasingly rely on the coordinated actions of multiple autonomous agents, each equipped with the ability to sense, decide, and act independently. These multi-agent systems have transformed numerous domains including robotics \cite{sony2020industry, liao2023optimization, poudel2021resource, poudel2023decentralized}, supply chain management \cite{giannakis2011multi, govindan2017supply, zhou2021multi}, communication networks \cite{su2019resource, gyory2024adaptation, chaudhari2022co}, transportation systems \cite{chamseddine2017bio, marwaha2015system, zhang2020multi}, and most recently, systems involving multiple Large Language Model agents \cite{wu2023autogen,lore2023strategic, chen2024instigating}. A fundamental challenge in these systems lies in their decentralized nature - while each agent operates based on its individual objectives and local information, the overall system must achieve broader collective goals. System managers face the complex task of steering these autonomous agents toward system-level objectives through strategic allocation of limited resources, such as communication channels or energy.

At the core of these decentralized systems lies the crucial element of inter-agent communication and interaction, which distinguishes multi-agent systems from independent single-agent scenarios. These interactions can be modeled as networks, with resources represented by network links. Research has shown that different network structures significantly impact system-level metrics including performance, resource utilization, and cooperation levels (See the background section). This raises a key question: How can we achieve desired system-level outcomes by strategically modifying these network structures?

The complexity of intervening at the network structure level is compounded by several factors inherent to decentralized multi-agent systems. The environment's dynamic nature, combined with agents' continuous learning and adaptation \cite{karcanias2010complexity}, makes developing effective coordination policies particularly challenging. The heterogeneity of agents - variations in properties, policies, and learning rates - further increases system complexity \cite{twu2014measure}. Additionally, the decentralized nature of these systems introduces partial observability, where both managers and agents must operate with limited information about the true environmental state \cite{jaakkola1994reinforcement}.

While Deep Reinforcement Learning \cite{lillicrap2015continuous, chen2021leveraging, liang2015state, shao2019survey, andriotis2019managing, chen2024sos, ororbia2024discrete} offers promising approaches for learning adaptable policies in dynamic, partially observable environments \cite{chen2022dynamic}, its application to network structure control faces a significant challenge: the vast discrete action space of possible network configurations. This space grows as $O(2^{N \times (N-1)/2})$ for a network with $N$ nodes, quickly becoming intractable for efficient policy learning as networks expand.

To address these challenges, we propose VAE-RL, a novel framework combining Variational Auto-Encoders (VAE) \cite{kingma2013auto} with Deep Reinforcement Learning. VAEs, comprising an encoder and decoder, can embed data distributions into latent spaces while maintaining the ability to reconstruct the original data. In our framework, we treat potential network configurations as a dataset, using a VAE to transform the vast discrete action space of network structures into a manageable continuous latent space. This enables the application of continuous-action DRL algorithms like Deep Deterministic Policy Gradient (DDPG), with the pre-trained VAE decoder reconstructing chosen latent actions into network structures.

Our work builds upon and extends our previous Two-tier framework for Systems of Systems (SoS) \cite{chen2022dynamic}, which introduced the concept of SoS workers (Tier I) and SoS managers (Tier II). While the previous framework was limited to homogeneous resource allocation, our current study enhances flexibility and efficiency by replacing the SoS manager with the VAE-RL framework. In this enhanced model, the manager controls communication network structures, with SoS workers sharing information through network connections. More connections indicate higher communication resource consumption, requiring the VAE-RL-equipped manager to balance system performance with resource usage through dynamic network modifications.

We evaluate our framework in a modified OpenAI Gym particle environment \cite{lowe2017multi}, testing both homogeneous and heterogeneous scenarios across various system scales. Our results demonstrate that VAE-RL outperforms baseline methods in optimizing the weighted sum of system-level performance and network-related resource usage, while revealing meaningful patterns for effective multi-agent system management.

\section{Background}
In this section, we review literature relevant to our research across three key areas. First, we explore network structure governance in various domains, which aligns with our paper's primary focus. Second, we examine generative models for network-based tasks, which relates to our approach of transforming discrete action spaces into continuous latent spaces. Finally, we review our previous Two-Tier framework, which serves as the foundation for evaluating our proposed VAE-RL method.

\subsection{Network Structure Governance}
Network structure governance - the ability to modify network configurations to influence agent behavior and system performance - has emerged as a critical research area across multiple domains. We examine its applications in three key areas: communication systems, public health policy, and socio-technical systems.
In communication systems, researchers have developed various approaches to network governance. Nakamura et al. \cite{Nakamura2022Virtual} address telecommunication service cost reduction while maintaining network stability under varying electricity prices. Bijami et al. \cite{Bijami2019A} introduce a Distributed Networked Control Scheme (DNCS) for stabilizing large-scale interconnected systems, specifically addressing communication challenges like random delays and packet dropouts. Deep Reinforcement Learning (DRL) applications in this domain include Chu et al.'s \cite{Chu2019Multi-Agent} scalable, decentralized Multi-Agent Reinforcement Learning (MARL) algorithm for adaptive traffic signal control, and Shibata et al.'s \cite{Shibata2021Deep} exploration of MARL for multi-agent cooperative transport. Network structures have also been studied in social and economic contexts, examining communication benefits, costs, and decentralized strategic decisions \cite{jackson2003strategic, heydari2015efficient}. These studies have identified structures that balance network stability and efficiency \cite{mosleh2016resource,mosleh2016distributed}.
Public health policy has increasingly embraced network governance approaches. Dynamic complex networks have proven effective in modeling micro-social interactions for public health \cite{cao2022micro}, urban policies \cite{ke2021airbnb}, and platform economies \cite{heydari2023reengineering}. During the COVID-19 pandemic, Mazzitello et al. \cite{Mazzitello2020Optimising} developed network-based pool testing strategies for low-resource settings, while Robins et al. \cite{ROBINS2023108} analyzed network interventions like social distancing. Siciliano et al. \cite{siciliano2021strategies} contributed a framework for network control in public administration to promote behavioral change.
In socio-technical systems, network governance has facilitated various advances. Ion et al. \cite{Ion2019A} developed a scalable algorithm for self-organizing control systems using networked multi-agent systems. Ellis et al. \cite{ellis2020implementing} studied network impacts on self-management support interventions, while Joseph et al. \cite{joseph2022qualitative} proposed a complex systems approach to model social media platform interventions. Network structure has proven crucial in promoting prosocial behaviors such as cooperation \cite{nowak2006five,gianetto2015network,gianetto2013catalysts}, coordination \cite{gianetto2016sparse}, and fairness \cite{mosleh2017fair}, particularly in social dilemma situations.
While network structure control has garnered significant attention, most existing approaches rely on traditional methods like hard-coded policies or heuristics, which struggle with complex, evolving, and partially observable environments. Deep Reinforcement Learning offers greater flexibility but often focuses on decentralized approaches, potentially compromising system-level optimization. Our approach addresses these limitations through a centralized method that tackles the curse of dimensionality.

\subsection{Generative Models for Networks}
Generative models have revolutionized various fields, including images \cite{goodfellow2020generative, creswell2018generative}, voice \cite{Kumari2022An, Tripti2022Image, Lakhotia2021On}, and text \cite{de2021survey, ray2023chatgpt}. In network modeling, Kipf and Welling \cite{kipf2016variational} introduced the Variational Graph Auto-Encoder (VGAE), enabling unsupervised learning on graph-structured data. Li et al. \cite{li2018learning} developed an approach using graph neural networks to express probabilistic dependencies among nodes and edges.
Domain-specific applications of network generation include DeepLigBuilder by Li et al. \cite{li2021structure} for drug design, Zhao et al.'s \cite{Zhao2021Physics} physics-aware crystal structure generation, and Singh et al.'s \cite{Singh2021Deep} TripletFit technique for network classification. Comprehensive surveys \cite{guo2022systematic, cheng2021molecular, navidan2021generative} provide extensive coverage of network generation methods.
Our work takes a novel approach by integrating generative models with Deep Reinforcement Learning to enable centralized network structure control while addressing the curse of dimensionality.

\subsection{Overview of the Two-Tier RL Governance Framework}
Our previous work \cite{chen2022dynamic} introduced a Two-Tier framework integrating Reinforcement Learning to optimize resource distribution in Systems of Systems (SoS). The framework operates across two training levels.
In Tier I, individual SoS components undergo decentralized-training-decentralized-execution (DTDE) using the Deep Deterministic Policy Gradient (DDPG) algorithm \cite{lillicrap2015continuous}. This tier develops essential skills based on partial observations, after which policies become fixed. Tier II focuses on the centralized resource manager's allocation decisions, including resource type and timing, using Deep Q-learning to maximize operational effectiveness.
While our previous work limited the system manager to choosing between empty or fully connected communication networks, our current research expands these capabilities. The manager can now assign any possible communication network configuration at each time step, enabling more adaptive and tailored strategies.

\section{Methodology}
In this section, we provide background knowledge and explain the method we are using. First, we briefly introduce the Partially Observable Markov Decision Process (POMDP), the fundamental model underlying Deep Reinforcement Learning (DRL). Next, we discuss the Deep Deterministic Policy Gradient (DDPG), the DRL algorithm we employ for managing continuous latent action spaces. We also explain the Branching Structured-Based DQN, an important baseline method in our study. Finally, we comprehensively introduce our methodology for training Variational Auto-Encoders (VAE) on networks and the integration of VAE with DDPG to optimize our system's performance.

\subsection{Partial observable Markov decision process}
The environmental dynamics in our study are aptly modeled using a Partial Observable Markov Decision Process (POMDP), as outlined in \cite{sutton2018reinforcement}. A POMDP is characteristically defined by a tuple $<S, {s^0}, {A}, {O}, T, {R}>$: where $S$ is a set of potential states; $s^0$ represents the initial state of the environment, with $s^0 \in S$; $A$ denotes the set of available actions; $O$ comprises the observations derived from the actual state; $R$ is the reward outputted by the environment; $T$ is the transition function, defined as $T:O \times A \rightarrow O$; and $U$ is the reward function, where $U:O \times A \rightarrow R$. The primary objective within a POMDP framework is to develop an optimal policy, $\pi_{\theta}$: $O \times A\rightarrow [0,1]$. This policy aims to maximize the expected return, calculated as $\sum_{t=0}^{n}\gamma^t{r}^t$, where $\gamma$ represents the discount factor.

\subsection{Deep deterministic policy gradient}
Policy gradient methods, as described in \cite{sutton2000policy}, represent a potent subclass of reinforcement learning (RL) algorithms. The fundamental concept involves directly modifying the policy parameters, denoted as $\theta$, to optimize the objective function $J(\theta) = E_{\pi_\theta}[R]$. This optimization is achieved by progressing in the direction of the gradient $\nabla_\theta J(\theta)$. The policy gradient can be expressed as follows: $\nabla_\theta J(\theta) = E_{o, a}[\nabla_{\theta}\log\pi_\theta(a|o)Q^\pi(o,a)]$.
Here, $Q^\pi(o,a)$, representing the action-value function, can be updated using techniques outlined in the Q learning section. This approach evolves into the actor-critic algorithm. Extending this framework to deterministic policies, denoted as $\mu_\theta:O \rightarrow A$, leads to the formulation of the Deep Deterministic Policy Gradient (DDPG) algorithm. In DDPG, the gradient of the objective undergoes a transformation: $\nabla_\theta J(\theta) = E_{o, a}[\nabla_{\theta}\mu_\theta(a|o)\nabla_a Q^\mu(o,a)|a=\mu_\theta(o)]$. In this context, $\mu_\theta(a|s)$, which represents the deterministic policy, is modeled using a deep neural network parameterized by $\theta$. The DDPG algorithm is particularly suited for scenarios involving continuous action spaces, as it outputs a deterministic action value at each time step.

\subsection{Branching Deep Q Network for Network-based system}
The Branching Deep Q Network (BDQN) \cite{tavakoli2018action} presents a solution for implementing Deep Q learning in scenarios characterized by large discrete action spaces. When an action space encompasses several dimensions, denoted as $N$, with each dimension offering a multitude of options, symbolized as $D$, the complexity of this space escalates exponentially, represented as $O(D^N)$. This exponential growth not only results in impractically large model sizes but also significantly complicates the optimization process. To address this, BDQN leverages the Dueling Q Network architecture \cite{wang2016dueling}, which separately learns Q functions for each dimension. It then integrates these functions using a shared state value, facilitating coordination among them. Consequently, this approach effectively reduces the action space complexity to $O(N*D)$, rendering the learning process feasible for tasks with extensive discrete action spaces.

To elucidate how BDQN serves as a baseline in our context, our action space of network topology can be converted into actions based on the number of link dimensions, where each dimension has two options — to have a link or not. Therefore, in our case, $N$ corresponds to the number of links, and $D$ is 2, representing the binary choice for each link (either present or absent). Utilizing the BDQN training scheme, we can effectively learn policies for network structure assignment.

\subsection{Variational Auto-encoder}
The Variational Auto-encoder (VAE) \cite{kingma2013auto}, a renowned generative model, has garnered considerable acclaim in fields like image and network generation. Its primary objective is to encode a dataset of independently and identically distributed samples into an embedded distribution that represents certain latent variables (encoder) and to subsequently generate new samples from this distribution (decoder). During the training phase, the encoder and decoder are trained jointly, facilitating a cohesive learning process. Upon establishing the embedded distribution, the decoder's parameters are fixed, enabling it to generate samples that mirror the distribution of the original dataset. This capability is the cornerstone of VAE's proficiency in producing images or networks akin to those in the dataset. In our specific context, the dataset comprises network topologies, represented by adjacency matrices. The training regimen adheres closely to the standard VAE protocol, which unfolds as follows:

In our dataset, let's denote the variables as $x$, with each data point following a distribution $p(x)$. The latent variables are characterized by the distribution $p(z)$. The encoder in a VAE is effectively represented by the conditional probability $p(z|x)$, while the decoder is represented by $p(x|z)$. Direct computation of $p(z|x)$ poses practical challenges, primarily because the underlying distribution $p(x)$ is intractable, as evidenced when expanding $p(z|x)$ using Bayes' rule. To circumvent this, VAE introduces an approximate function $q(z|x)$, assuming its tractability. In traditional VAE models, a standard multivariate Gaussian distribution is employed for this approximation. To refine our approximation of $p(z|x)$, we focus on minimizing the Kullback–Leibler divergence between $q(z|x)$ and $p(z)$. This minimization process involves a series of mathematical operations, ultimately leading us to maximize the following objective: $E_{q(z|x)}\log p(x|z) - \text{KL}(q(z|x) || p(z))$.

The objective function in VAEs comprises two key components. The first is the log-likelihood of the reconstructed variables, which measures the accuracy of the reconstruction. The second component is a regularization term, which ensures that the distribution of the latent variable $z$ aligns closely with the prior distribution. As previously mentioned, this prior distribution is assumed to be a standard multivariate Gaussian distribution. To model both the encoder $q(z|x)$ and decoder $p(x|z)$, deep neural networks are employed. Practically, the encoder outputs the mean and covariance of the Gaussian distribution from which $z$ is sampled. Furthermore, VAE employs reparameterization tricks to enable the backpropagation of gradients through stochastic nodes, thereby facilitating effective training of the network. For a more comprehensive understanding of these mechanisms, further details are available in \cite{kingma2013auto}.

\begin{figure*}
\includegraphics[scale = 0.5]{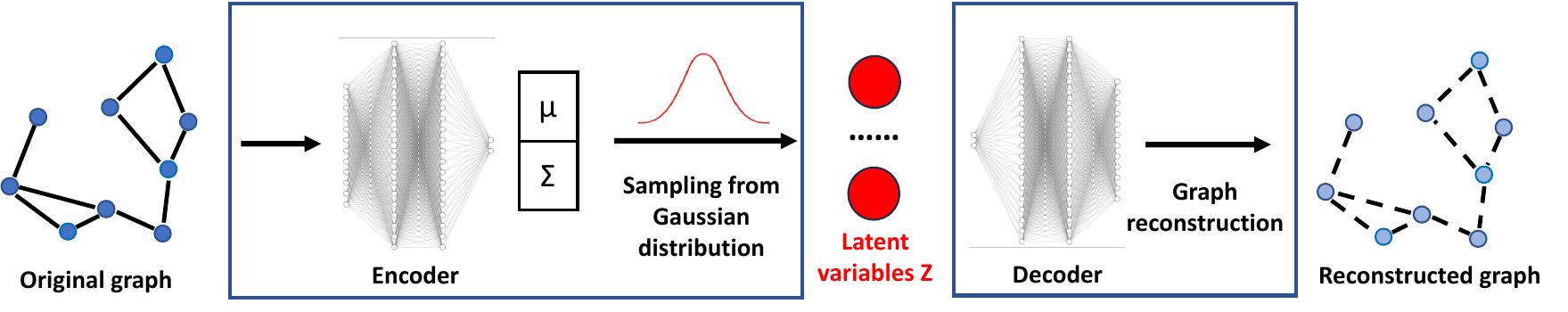}
\centering
\caption{The diagram shows a Variational Autoencoder applied to network topology. The encoder processes the adjacency matrix, producing Gaussian distribution parameters. The decoder samples from this distribution to reconstruct the adjacency matrix. Both components use deep neural networks.}
\label{fig:VAE_training}
\end{figure*}

\subsection{DDPG with VAE for network topology governance}

In controlling multi-agent systems at the system level, a centralized management approach is pivotal, primarily to harmonize high-level performance with the allocation of various types of resources. This approach's efficacy is substantiated in studies such as \cite{chen2022dynamic, chen2024sos}. Within these complex systems, the dynamics of information sharing and agent interaction are crucial. The topology of the network, in particular, plays a crucial role in influencing outcomes across different scenarios. Consequently, when controlling multi-agent systems, a frequent challenge encountered is the need for centralized control over the network topology. This task is complicated by the sheer complexity and diversity of potential network topologies, resulting in an action space that is vast and often impractical for traditional control algorithms with discrete action spaces, such as Deep Q-learning \cite{mnih2015human}.

Policy gradient methods are notably proficient in handling complex and continuous action spaces \cite{lillicrap2015continuous}. In our cases, where the inherent action space comprises network topologies, it will be beneficial to encode this vast, discrete action space into a manageable, continuous latent action space. By applying policy gradient methods to this transformed, latent continuous space, we can adeptly navigate and optimize within it. Subsequently, the optimized latent space can be decoded back into reconstructed network topologies. This approach effectively circumvents the challenges associated with the original, large discrete action space, presenting a more efficient method for network topology construction and optimization.

Inspired by this idea, we introduce the VAE-RL framework as a solution. This approach begins by employing a Variational Auto-encoder (VAE) to transform the extensive, discrete action space associated with network topologies into a continuous action space. Following this transformation, policy gradient algorithms with continuous action spaces are utilized to learn effective control policies. A key advantage of this framework is its ability to leverage the generative capabilities of the VAE, allowing for the reversion of the controlled, continuous action space back to its original form — the network topology. This process effectively surmounts the challenge posed by the vast discrete action space, thereby enhancing the control mechanisms within multi-agent systems. The training process for the VAE-RL framework is separated into two distinct phases: the initial training of the VAE, followed by the DRL training, which is conducted using the pre-trained VAE.

\subsubsection{VAE training process}

Our primary objective centers on managing network topologies within multi-agent systems. To achieve this, we concentrate on training a Variational Auto-encoder (VAE) specifically designed to encode adjacency matrices into a latent space, and then decode this latent space back into reconstructed adjacency matrices. For both the encoder and decoder components of the VAE, we utilize fully connected neural networks. However, in scenarios where the system manager possesses comprehensive control with higher authority, extending not only to network topology but also directly to agent properties, Graph Neural Networks (GNNs) \cite{zhou2020graph} emerge as a viable alternative for representing both the encoder and decoder. The training framework for the VAE, as applied to network topology, is depicted in Figure \ref{fig:VAE_training}.

Initially, we assemble a dataset, denoted as $D$, comprising samples of potential network topologies given the number of nodes. It's crucial to acknowledge that encompassing every possible network topology becomes more intractable as the number of nodes increases. Subsequently, this dataset is partitioned into a training set and a validation set, adhering to conventional supervised learning methodologies. Network topologies are represented through a flattened adjacency matrix, symbolized as $A$.

The architecture of our model employs fully connected neural networks to instantiate the encoder and decoder, denoted by $f_{encoder}(x;\theta_1)$ and $f_{decoder}(x;\theta_2)$, respectively. During each training iteration, a mini-batch of data $a$ is sampled from the training set. The encoder processes this data to yield the mean and covariance of a multi-variate Gaussian distribution: $\mu, \Sigma = f_{encoder}(a;\theta_1)$. Subsequently, the latent variable values $z$ are sampled from this distribution, where $z\in R^d$ and $d$, the latent variable dimension, is a pre-defined hyperparameter.

The decoder then takes $z$ as input and produces the reconstructed adjacency matrix $\hat{a}$: $\hat{a} = f_{decoder}(z;\theta_2)$. Loss is calculated using the standard VAE loss we mentioned earlier, and the gradient of the loss is backpropagated to update $\theta_1$ and $\theta_2$ in $f_{encoder}(x;\theta_1)$ and $f_{decoder}(x;\theta_2)$. Ultimately, the model is evaluated on the validation set, and the model with better performance through validation is saved.

\subsubsection{DDPG training process with learned VAE}
\begin{figure}
\includegraphics[scale = 0.45]{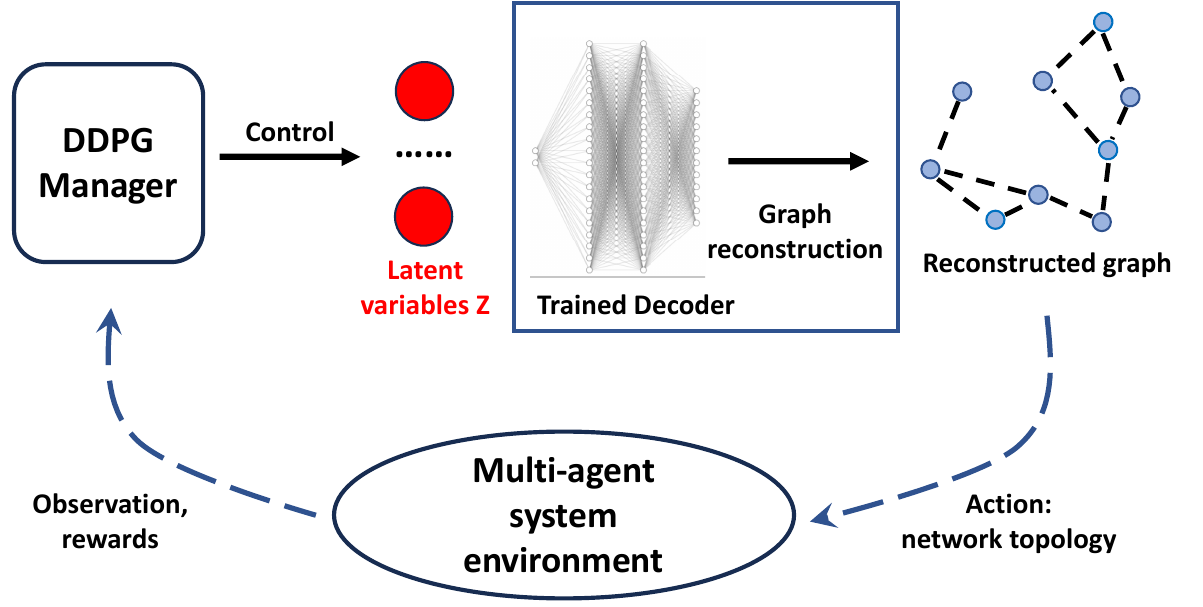}
\centering
\caption{The diagram depicts the VAE-RL framework used for interaction within a multi-agent system environment. The DDPG manager directly manages the continuous latent variables, which are decoded by the decoder into the reconstructed network topology. This topology serves as the final action within a Partially Observable Markov Decision Process (POMDP). Following this, the multi-agent system environment processes the action and updates its state accordingly. The DDPG Manager then receives observations and rewards, which it uses to update its policy and execute subsequent controls. During this process, the decoder's parameters remain fixed from the pre-trained model. }
\label{fig:VAE-RL-diagram}
\end{figure}

In the VAE training phase, we have two key outcomes: a learned encoder that effectively embeds input network topologies into a latent space, and a learned decoder capable of reconstructing network topologies from latent variables. During the DDPG training, the parameters of the pre-trained VAE are fixed, with an exclusive focus on utilizing the decoder. This process is illustrated in Figure \ref{fig:VAE-RL-diagram}.

The environment dynamics are defined as $R, S' = Transition(S, A_{adj})$, where $R$ represents the reward for the current step, $S$ is the current state, and $S'$ is the subsequent state of the environment. The current action, $A_{adj}$, corresponds to the network topology. For modeling the actor $\mu(o;\phi)$ and the critic $Q(o, a;\theta)$, fully connected networks are employed. In each step of the DDPG training, the manager observes the current state of the environment, obtaining the observation $o^t$ at time step $t$. The actor then outputs latent variable values: $z^t = \mu(o^t;\phi)$. As part of exploration, noise is added to actions, diminishing over time, a strategy recommended by \cite{lillicrap2015continuous}. Subsequently, the decoder transforms these latent variables into a reconstructed adjacency matrix: $a_{adj}^t = f_{decoder}(z^t; \theta_2)$. The environment takes this reconstructed matrix, updates its state, and generates corresponding rewards: $r, s^{(t+1)} = Transition(s^t, a_{adj}^t)$. This trajectory, denoted as $s^t, z^t, r, s^{(t+1)}$, is stored in the Replay-buffer $D$. During training, a mini-batch of trajectory data is sampled; the actor and critic are updated by backpropagating the gradient of the loss in the DDPG we mentioned earlier in Section 3.2.

\section{Experiment results and discussion}

In this section, we first introduce the environment we are using and the related settings. Next, we compare the performance of the proposed VAE-RL framework to baseline methods in different scenarios. We then analyze the learned behavior of the system manager using VAE-RL and discuss the trends and insights derived from the results. Finally, we present snapshots of the network evolution over time, visualizing the system's dynamics and justifying our findings.

\subsection{Experiment design}

We utilize the modified OpenAI Gym particle environment \cite{lowe2017multi} to evaluate the effectiveness of our VAE-RL framework. In the original 'spread' task within this particle environment, multiple agents are tasked with spreading themselves on landmarks while minimizing collisions. This scenario constitutes a multi-agent system where effective coordination among agents is essential. In our previous work \cite{chen2022dynamic}, we introduced modifications to this environment, making it partially observable. Furthermore, we introduced different types of resources, such as additional vision and communication capabilities. These resources are dynamically selected by the SoS manager to maintain a balance between system performance and resource utilization for the SoS workers. However, the communication network options were limited (either empty or complete) in the previous study. Additionally, the SoS agents were homogeneous, and the proposed framework fell short in handling heterogeneous situations.

In this paper, we focus solely on the allocation of communication resources and adding flexibility to the action space. The SoS manager can select different communication network topologies during tasks, providing opportunities to save communication resources and improve system performance in uncertain environments with heterogeneous agents. Therefore, we retain the essential components in the hierarchical framework for SoS that we previously proposed but mainly increase the flexibility of network topologies within the action space in Tier II.

In Tier I, SoS workers are initialized in a manner consistent with the previous framework. SoS workers and landmarks are randomly initialized at the beginning of each game within a $2 \times 2$ square. They possess limited vision, allowing them to observe only landmarks or other agents within their visual range. Furthermore, these workers act autonomously, making individual decisions based on the information available to them. While the SoS manager in Tier II lacks the ability to directly compel SoS workers to alter their actions, it can influence their behavior by manipulating the information they receive. This manipulation is achieved through the assignment of different communication network topologies. It's important to note that the SoS manager does not possess an all-encompassing view of the entire system but instead relies on aggregate information derived from the observations of SoS workers.

At each time step, the SoS manager assigns a communication network to the SoS workers. Subsequently, these workers share their information with their neighbors within the communication network. The allocation of communication networks involves the utilization of communication resources, with each link in the network incurring a cost of 0.1 in the current settings. The primary objective of the SoS manager is to maximize a weighted combination of two key sub-objectives: maximizing task performance and minimizing communication resource costs. The current weight is 1, which means the scores of tasks and the communication resource cost are equally important. The specific weights applied to these sub-objectives can be adjusted based on their relative importance, which is thoroughly explored in \cite{chen2022dynamic}.

Regarding the VAE-RL algorithms, we set several hyperparameters: the decoder and encoder in VAE, and the actors and critics in DDPG are represented using fully connected neural networks. For VAE, we use 512 $\times$ 256 to represent the encoder and 256 $\times$ 512 to represent the decoder, while the dimension of the latent space is 10 for the multi-agent system with 10 agents. The resource manager uses 1024 $\times$ 512 $\times$ 256 fully connected neural networks to represent critics and 1024 $\times$ 512 fully connected neural networks to represent actors. We use the Adam optimizer with a 0.001 learning rate for critic training and the Adam optimizer with a 0.0001 learning rate for actor training. The training for resource managers includes 20,000 epochs. It is crucial to note that critics are only used during the training phase, so only actors participate in the decision-making process during the execution phase. In our experiments, each game has 50 time steps, and the following results have been tested on 1,000 new games.

\begin{figure}
\includegraphics[scale = 0.5]{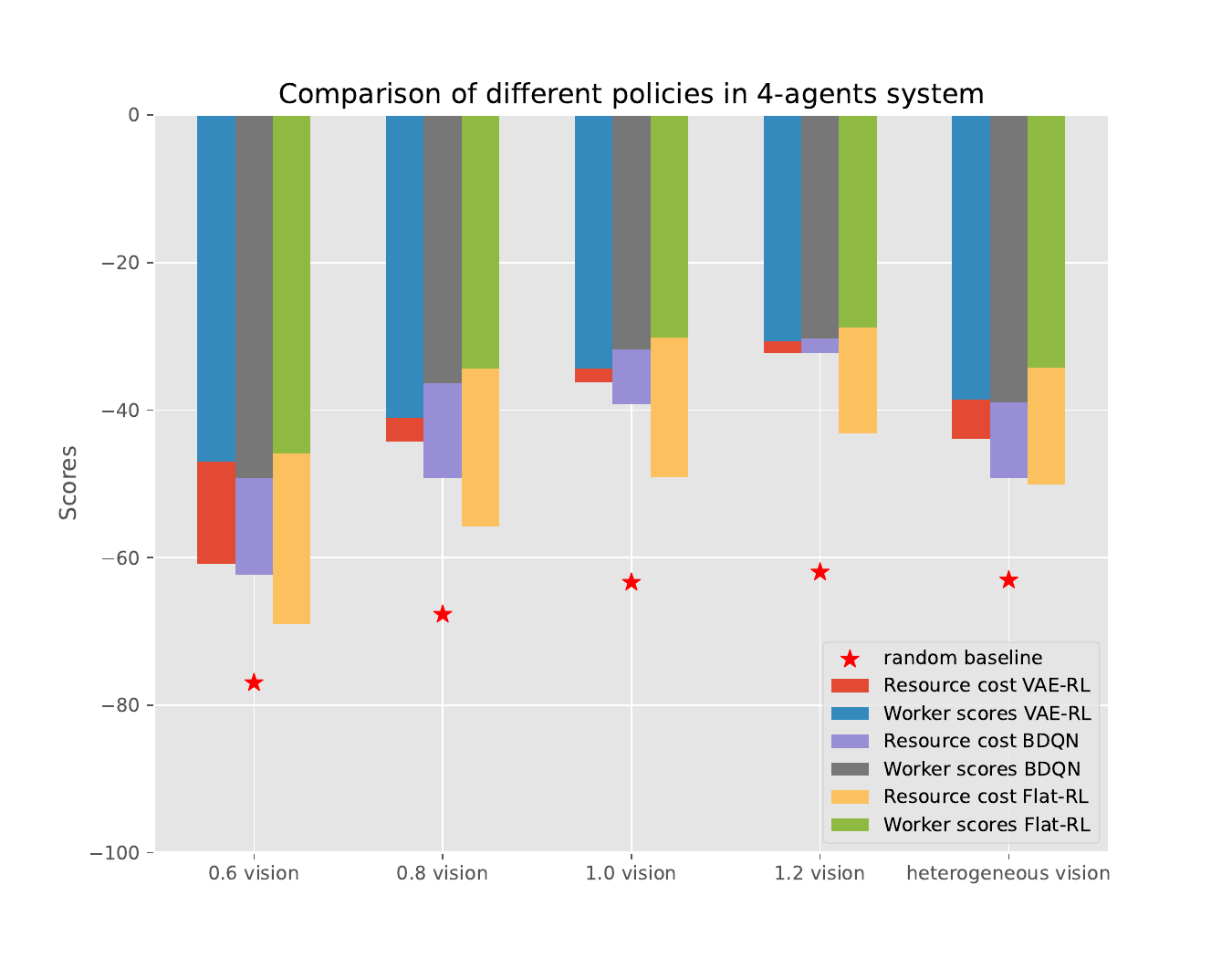}
\centering
\caption{Results show performance and resource penalties for various methods with homogeneous agents (vision ranges 0.6-1.2) and heterogeneous agents in 4-agent systems. A star marker indicates the random policy baseline's overall performance.}
\label{fig:performance_4agents}
\end{figure}

\subsection{Performance of proposed method}
To assess the generalization and robustness capabilities of our proposed VAE-RL method, we aim to compare its performance against the traditional DRL approach using a discrete action space for network topology, hereafter referred to as Flat-RL, within a 10-agent system. Given that the action space in this system is $O(2^{N*(N-1)})$, where $N=10$, its enormity renders the training for Flat-RL impractical. Therefore, we introduce another baseline, BDQN \cite{tavakoli2018action}, which is scalable in larger networks. Initially, we apply both VAE-RL and the two baseline methods (Flat-RL and BDQN) to a smaller system comprising 4 agents. After establishing that BDQN demonstrates equivalent or superior performance compared to Flat-RL in this smaller setup, we then extend our evaluation to a larger 10-agent system. Here, BDQN's performance serves as a proxy for the upper bound of Flat-RL's capabilities. This approach allows us to indirectly gauge the performance of Flat-RL in larger systems, thereby ensuring the coherence of our results.

Initially, we established a homogeneous vision range for all SoS workers. We designed a series of four experiments, varying in difficulty levels, with vision ranges of 0.6, 0.8, 1.0, and 1.2. It's worth noting that all entities are positioned within a $2 \times 2$ square configuration, meaning that even the simplest task remains partially observable. Additionally, even tasks with a vision range of 1.2 are not trivial because it is still possible that agents are unable to observe anything at the initial states. In numerous real-world multi-agent systems, agents often exhibit heterogeneity in their properties or capabilities. To test our framework in a more realistic environment, we have devised an environment with heterogeneous agents, where agents have vision ranges of 2, 1.5, 1, and 0.5. These experiments serve as crucial and representative examples within the realm of heterogeneous scenarios.

\begin{figure}[t]
\includegraphics[scale = 0.5]{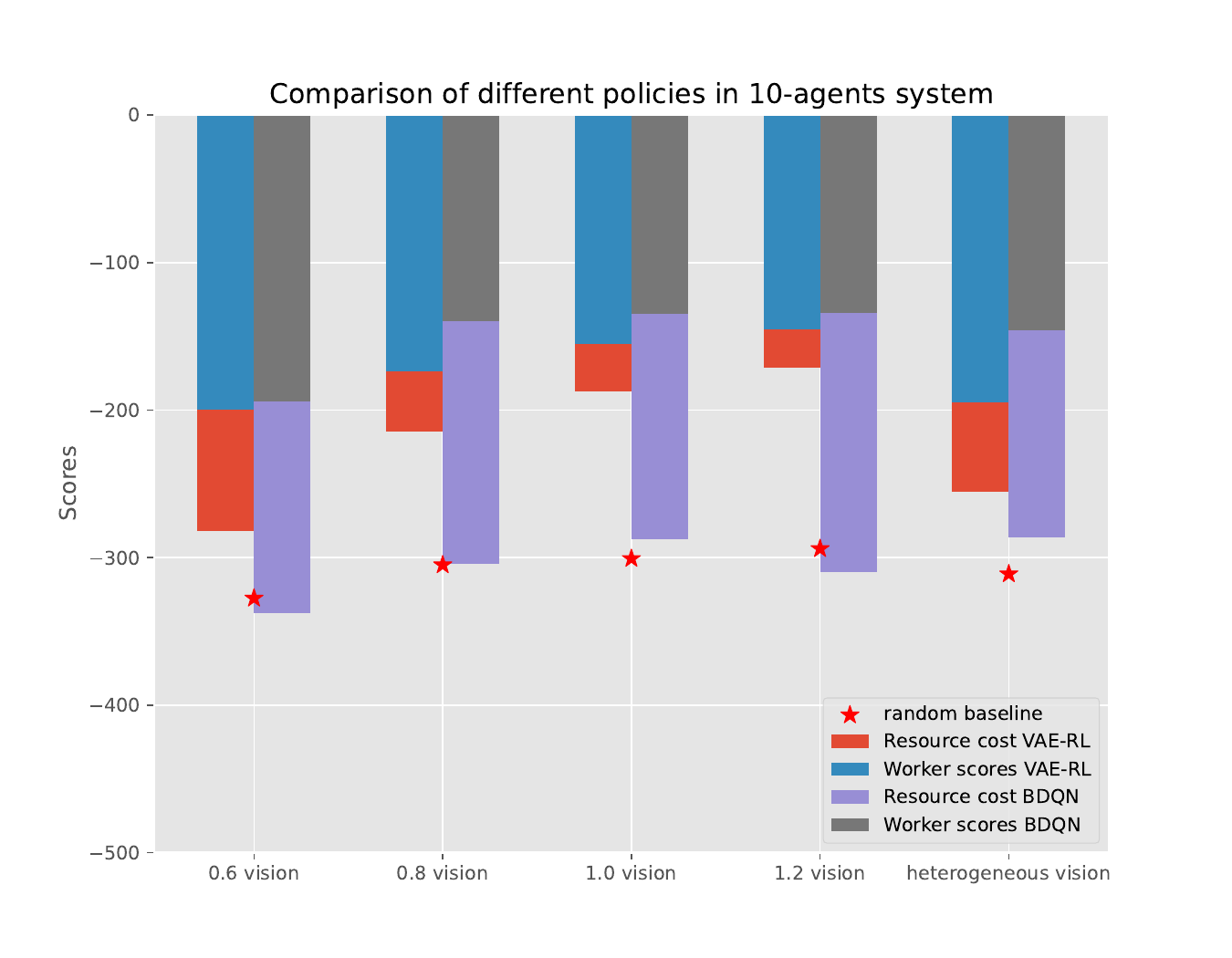}
\centering
\caption{Results show performance and resource penalties for various methods with homogeneous agents (vision ranges 0.6-1.2) and heterogeneous agents in 10-agent systems. A star marker indicates the random policy baseline's overall performance.}
\label{fig:performance_10agents}
\end{figure}

In our experiments, we first assess the performance of our VAE-RL framework, followed by an analysis of the learned behaviors of the Systems of Systems (SoS) manager. It is important to note that since the VAE training employs a standard supervised scheme with distinct training and validation sets within a sampled dataset, both the VAE-RL framework and its training process can be scaled to larger networks using the same training scheme by increasing the dataset and the neural networks to reasonable sizes.

\subsubsection{Results on two environments}
We first applied all methods on the small environment with 4 agents and got the results in Figure \ref{fig:performance_4agents}. From the results, our VAE-RL approach consistently outperforms other baseline methods across all tasks including heterogeneous cases, ranging from difficult to easy. This not only underscores the superior performance of our method but also underscores its robustness under different scenarios. Because it is hard to compare the homogeneous scenarios and heterogeneous scenarios directly, for the purpose of analyzing trends, our subsequent findings and discussions in this section will primarily focus on homogeneous cases.

\begin{figure*}[t]
    \centering
    \begin{subfigure}{.4\textwidth}
        \centering
        \includegraphics[width=\linewidth]{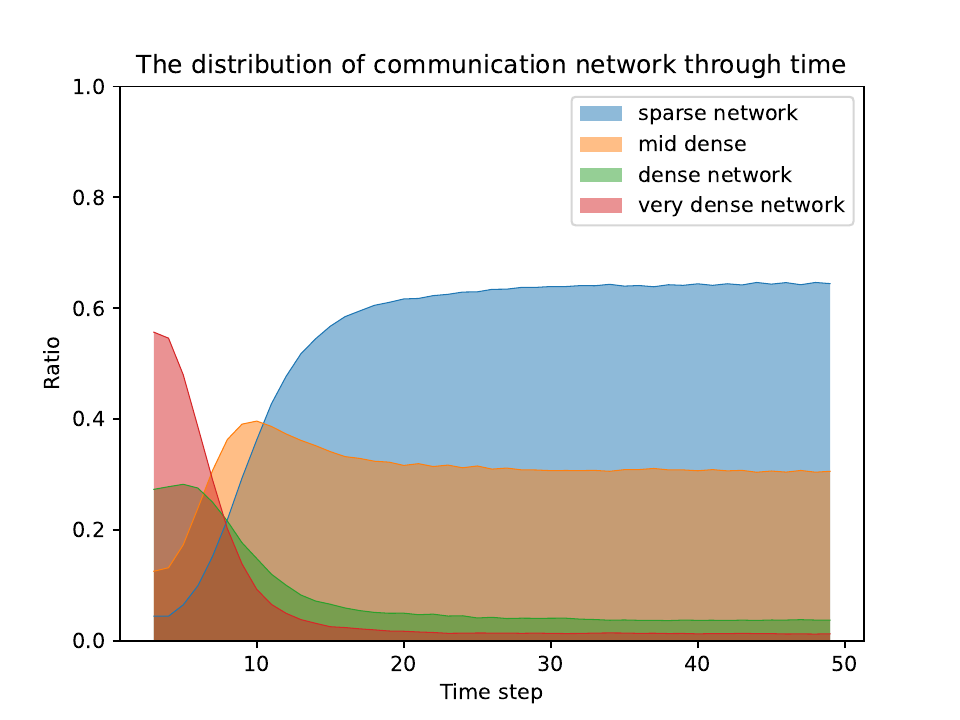} 
        \caption{0.6 vision range}
    \end{subfigure}%
    \hspace{2pt}
    \begin{subfigure}{.4\textwidth}
        \centering
        \includegraphics[width=\linewidth]{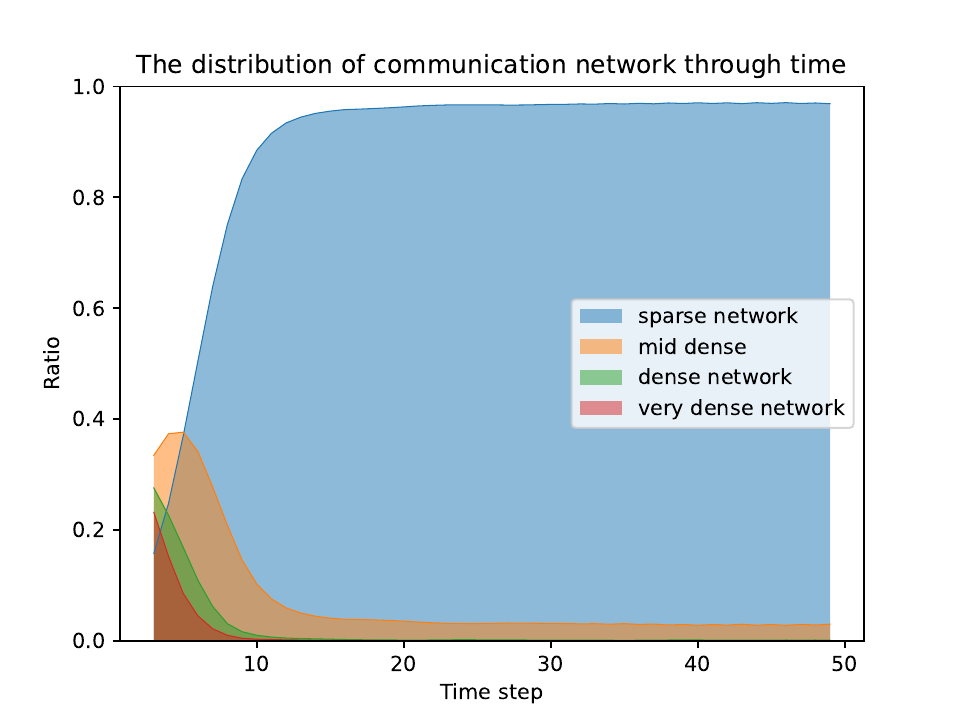} 
        \caption{0.8 vision range}
    \end{subfigure}%

    \begin{subfigure}{.4\textwidth}
        \centering
        \includegraphics[width=\linewidth]{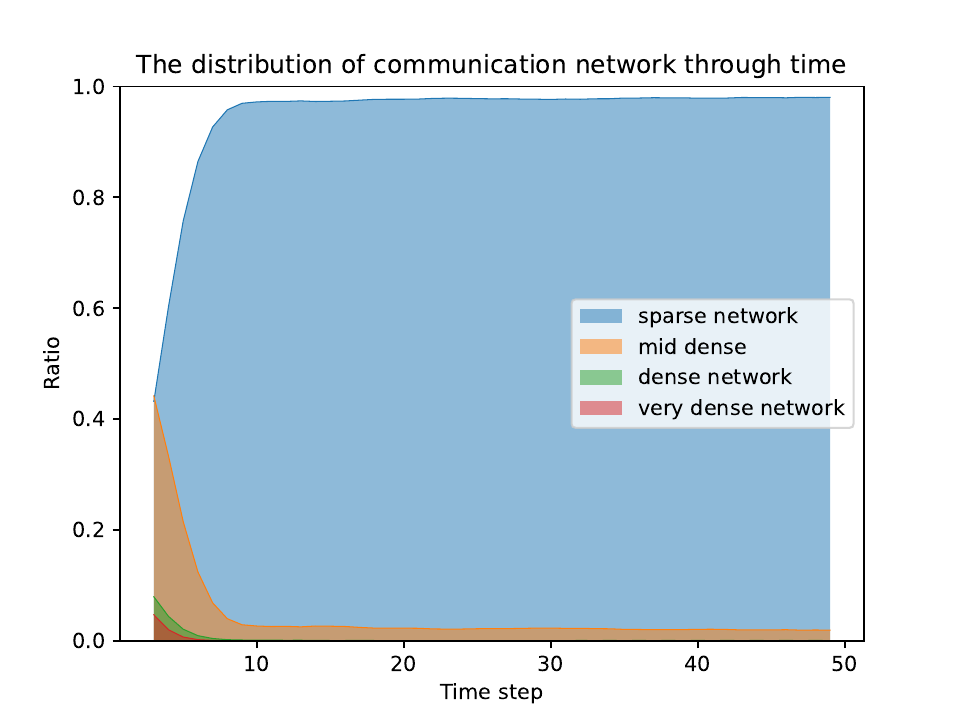} 
        \caption{1.0vision range}
    \end{subfigure}%
    \hspace{2pt}
    \begin{subfigure}{.4\textwidth}
        \centering
        \includegraphics[width=\linewidth]{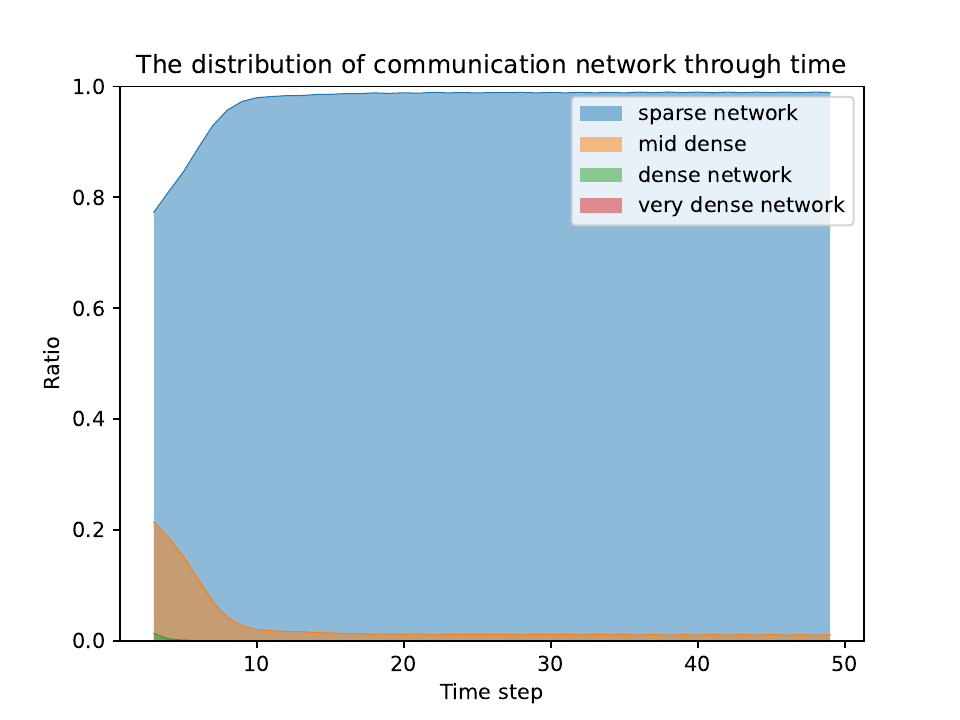} 
        \caption{1.2 vision range}
    \end{subfigure}

    \caption{Communication network distribution over time is analyzed for homogeneous agents with vision ranges of 0.6, 0.8, 1.0, and 1.2 (subgraphs a-d). Networks are categorized as sparse ($\le$ 9 links), mid-dense (9-17 links), dense (18-26 links), or very dense ($\ge$ 27 links).}

    \label{fig:behavior_homo}

\end{figure*}

Secondly, it's noteworthy that as the vision range increases, the performance of all methods and communication resource usage demonstrate improvement. This phenomenon can be attributed to the tasks becoming progressively easier. Even the simplest strategy, such as randomly selecting actions, becomes more effective because SoS workers have access to more self-observed information with larger vision ranges, leading to enhanced performance. Furthermore, with larger vision ranges, workers have increased opportunities to independently observe landmarks and other agents. Consequently, the reliance on the communication network diminishes, leading to reduced usage of communication resources.

Thirdly, as tasks become easier, the VAE-RL method exhibits substantial improvements compared to the baseline methods. In environments where SoS workers have limited vision ranges, such as 0.6 and 0.8, the tasks are exceedingly challenging for agents to accomplish with flawless performance, even when employing intelligent communication network assignment strategies. This difficulty arises from the fact that smaller vision ranges result in SoS workers having very limited opportunities for useful observation at initial states. Despite the potential for communication through the network, they have little to share due to the scarcity of information. Consequently, in these critical scenarios, we observe only marginal enhancements in the VAE-RL method compared to the baseline methods. Conversely, when tasks are moderately challenging, the VAE-RL method exhibits more substantial improvements. This is because it leverages knowledge gleaned from the diverse positions of agents and landmarks to intelligently select communication networks that optimize system performance while conserving communication resources.

Finally, it becomes evident that although BDQN does not surpass VAE-RL in total performance, it consistently outperforms Flat-RL across a variety of scenarios, including those involving heterogeneous cases. This consistent superiority of BDQN over Flat-RL serves as a compelling indicator of its effectiveness, suggesting that BDQN's performance is on par with, or perhaps even superior to, Flat-RL. When considering larger and more complex environments that incorporate a greater number of agents, where the application of Flat-RL is hindered by the model's size and optimization complexities, BDQN stands as a viable baseline. Its proven advantage over Flat-RL in smaller settings provides a rationale for employing BDQN as a benchmark in these larger scenarios. This analysis allows for meaningful comparison and evaluation of the VAE-RL's performance against a relevant and established baseline in more expansive and challenging environments.

In our expanded experiment involving a larger environment with 10 agents, we tested the VAE-RL method and the BDQN baseline. As previously noted, Flat-RL is not capable of handling this more complex task, leading us to focus solely on the performances of VAE-RL and BDQN, as illustrated in Figure \ref{fig:performance_10agents}. The results clearly show that VAE-RL continues to outperform the BDQN baseline across all environment settings, including those with heterogeneous conditions. This consistent pattern reinforces the trends observed in smaller environments, indicating that VAE-RL's effectiveness scales well with increased complexity. On the other hand, the BDQN baseline struggles even in the simplest case with agents having 1.2 vision ranges, where its performance is comparable to a random strategy. This outcome suggests that while BDQN remains applicable in this context, it fails to develop a meaningful policy for communication network assignment. In contrast, VAE-RL not only demonstrates strong performance relative to the baseline but also retains its potential for effective implementation in even larger systems with more agents.

\begin{figure*}
    \centering
    \begin{subfigure}{.4\textwidth}
        \centering
        \includegraphics[width=\linewidth]{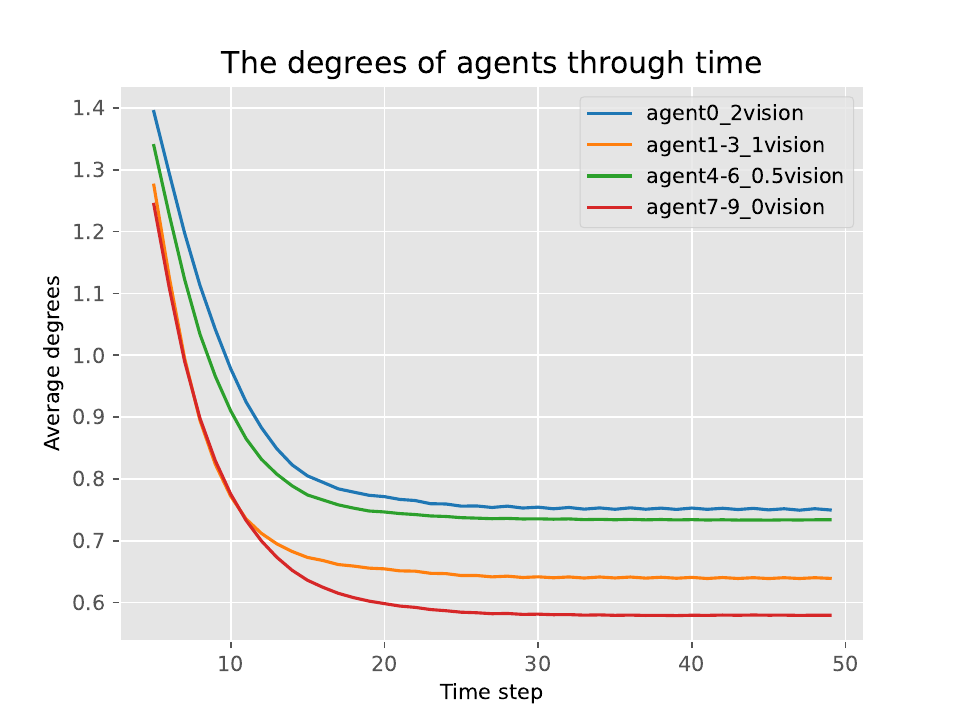} 
        \caption{Evolution of node degrees}
    \end{subfigure}%
    \hspace{0.5cm}
    \begin{subfigure}{.4\textwidth}
        \centering
        \includegraphics[width=\linewidth]{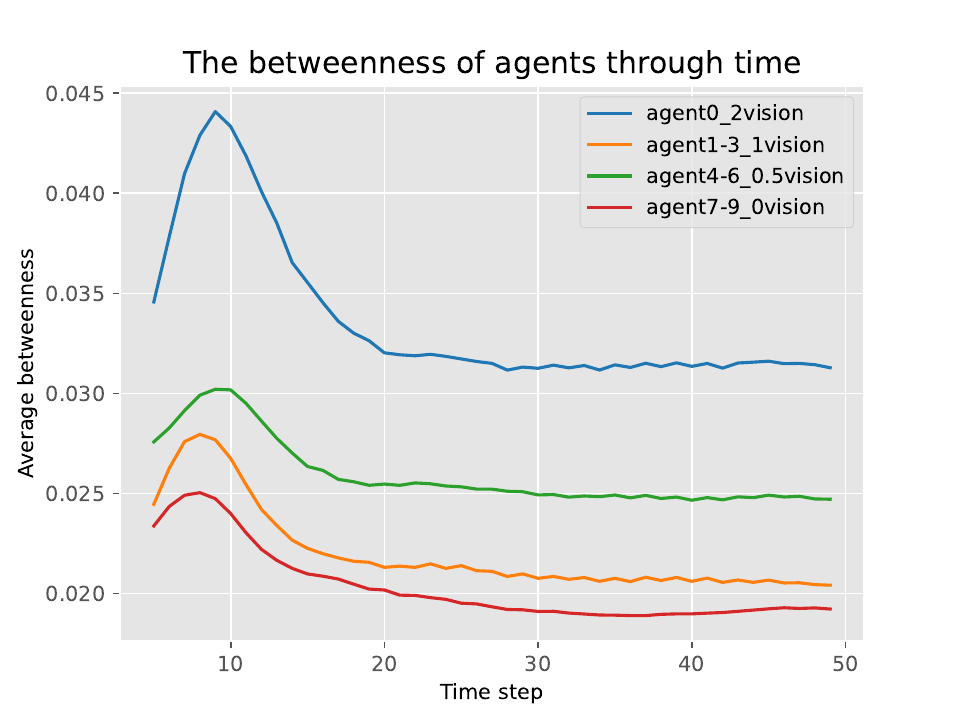} 
        \caption{Evolution of node betweenness}
    \end{subfigure}%
    \caption{The results show average node degrees and betweenness centrality for agents with different vision ranges over time. The study includes one agent with vision 2.0, and three agents each with visions 1.0, 0.5, and 0.}
    \label{fig:network_analysis}

\end{figure*}
\subsection{Evolution of Network Behavior}
After confirming that our VAE-RL method outperforms the baseline in both small and large systems, we also aim to analyze the learned behaviors of SoS managers to enhance the explainability of our models using deep learning techniques and generate valuable insights or heuristics. We illustrate the distribution of communication networks with varying link densities during the task. We examine the behaviors of VAE-RL's policy in both homogeneous and heterogeneous cases.

\subsubsection{Scenarios with homogeneous agents}

We have four environments for homogeneous cases, where agents have vision ranges of 0.6, 0.8, 1.0, and 1.2. We categorize the communication networks based on the number of links they contain. Sparse networks are defined as those with fewer than 9 links; mid-dense networks as those with more than 9 but fewer than 18 links; dense networks as those with more than 18 but fewer than 27 links; and very dense networks as those with more than 27 links. We then plot the distribution of networks with varying densities over time in Figure \ref{fig:behavior_homo}.

Firstly, the graphs show that as tasks become less challenging, there is a notable increase in the frequency of using less dense communication networks, while the frequency of employing costly denser communication networks decreases. In the easiest task, with a vision range of 1.2, the use of sparse communication networks approaches approximately 100\% at the end of the game. This observation aligns with our earlier explanation that easier tasks correspond to reduced utilization of communication resources.

Secondly, the behavior of the SoS manager exhibits two distinct phases: during the initial 10 time steps, the manager leans towards employing more costly communication networks with a higher number of links, then it shifts towards utilizing cheaper communication networks. At the beginning of each task, due to the constraints posed by limited vision ranges, some SoS workers may not have the capacity to observe any entities. Consequently, the manager's preference is to encourage all workers to aggregate their information through dense communication networks. However, denser communication networks entail higher communication resource costs. Therefore, as workers approach landmarks and can perceive landmarks within their own limited vision ranges, the manager is inclined to assign sparser communication networks to save communication resources. For instance, in the initial phase of sub-graph (a), where the vision range is 0.6, resulting in exceptionally challenging tasks, the manager exhibits a frequency of assigning very dense communication networks of almost 60\%. Subsequently, there is an immediate decrease in the frequency of employing costly communication networks, while the frequency of employing sparse communication networks exceeds 50\% shortly after a few time steps.

\subsubsection{Extending the model to heterogeneous agents}

In numerous real-world multi-agent systems, agents often exhibit heterogeneity in their properties or capabilities. Regrettably, many studies in the domain of multi-agent system control or SoS control \cite{chen2022dynamic} tend to exclusively focus on homogeneous scenarios, rendering them less adaptable for deployment in heterogeneous environments. To address this limitation, we have designed an experiment with 10 agents where agents possess heterogeneous vision ranges: (2, 1, 1, 1, 0.5, 0.5, 0.5, 0, 0, 0). This experiment serves as a crucial and representative example within the realm of heterogeneous scenarios. Instead of analyzing the communication networks in general, we focus more on the agents' properties within networks. Thus, we examine the average node degrees of the agents and the average node betweenness centrality for those with the same vision range and plot these through time. The calculation of node betweenness centrality includes both the start nodes and end nodes. Our analysis of the learned behaviors of the resource manager and the corresponding results are depicted in Figure \ref{fig:network_analysis}.

Upon analysis, we can discern two distinct phases in the behavior patterns of the SoS manager, paralleling our previous findings. During the first phase (0-10 time steps), the node degrees of agents are relatively high, indicating a greater level of connectivity among agents. This is quickly followed by a shift to a state with fewer degrees. In the second phase (10-50 time steps), a notable stabilization occurs, with the agents maintaining generally low node degrees. The rationale behind this pattern mirrors our earlier analysis: At the beginning, agents require a higher degree of information exchange for effective coordination and task completion, leading to an increased number of connections. However, maintaining a high node degree comes with substantial costs. Consequently, once the agents have acquired essential information and start converging toward their objectives, there's a significant reduction in node degrees to minimize costs while maintaining efficiency.

For node betweenness centrality, we observe an increase during the 0-10 time steps, a decrease between 10-20 time steps, and stabilization afterward. Since betweenness centrality measures the importance and effectiveness of each node in the information flow, we infer that initially, agents cannot effectively share their information due to their random positions and connections within the network. With guidance from the VAE-RL manager, agents are strategically positioned within more reasonable network structures, enhancing their importance and effectiveness in the network. Finally, as agents approach their targets, the need for a complex communication network diminishes, reducing the overall importance of all agents and resulting in stabilization.

Furthermore, we sought to identify patterns among agents with differing vision ranges. It was observed that agents with a vision range of 2 consistently exhibited the highest node degree and node betweenness centrality over time. This suggests that agents with a vision range of 2 tend to remain central within the communication network, maintaining more connections and a higher level of importance. Contrary to expectations, a clear trend where agents with greater vision ranges demonstrate larger node degrees and node betweenness centrality was not evident. Specifically, agents with a vision range of 1 generally had lower node degrees and node betweenness centrality compared to agents with a vision range of 0.5.

We propose several explanations for our findings. First, agents with moderate abilities serve a dual role: they require assistance from others to solve tasks in certain scenarios, yet they are also capable of providing valuable aid in different contexts. This bidirectional flow of information in a complex environment may contribute to the observed phenomenon. Second, while agents with moderate abilities lack the power to significantly assist others, they are able to solve tasks independently. This suggests that in some cases, they should be isolated from the communication network, resulting in lower node degrees and betweenness centrality compared to agents with lesser abilities. More details can be found in Appendix B.

\begin{figure}[t]
    \centering
    \begin{subfigure}{.2\textwidth}
        \centering
        \includegraphics[width=\linewidth]{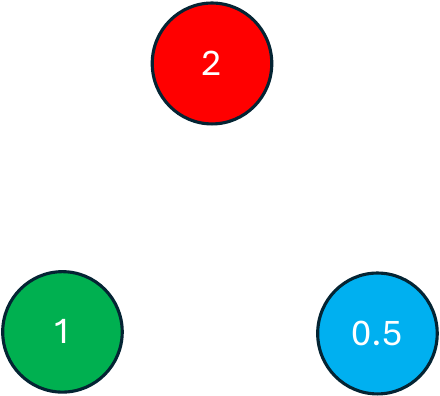} 
        \caption{Requirement of 0.5 ability}
    \end{subfigure}%
    \hfil
    \begin{subfigure}{.2\textwidth}
        \centering
        \includegraphics[width=\linewidth]{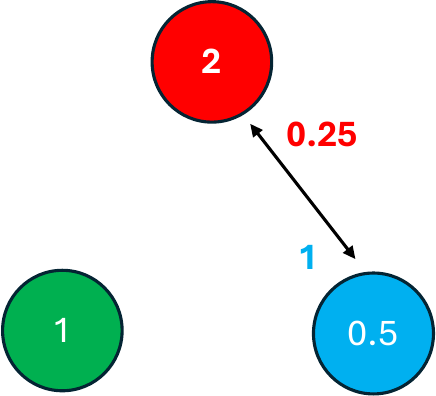} 
        \caption{Requirement of 1 ability}
    \end{subfigure}%
    
    \vspace{5mm}    
    
    \begin{subfigure}{.2\textwidth}
        \centering
        \includegraphics[width=\linewidth]{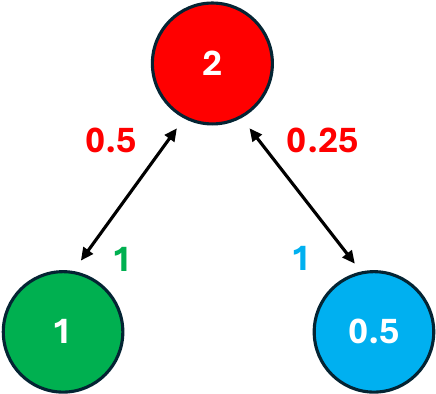} 
        \caption{Requirement of 1.5 ability}
    \end{subfigure}%
    \hfil
    \begin{subfigure}{.2\textwidth}
        \centering
        \includegraphics[width=\linewidth]{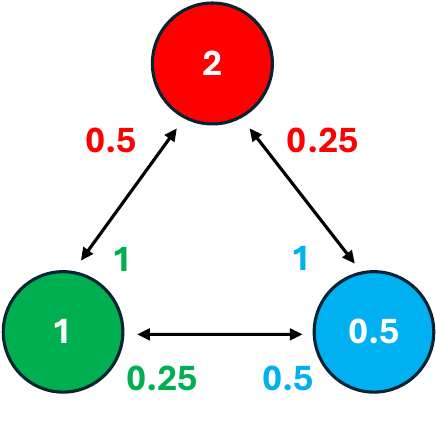} 
        \caption{Requirement of 2 ability}
    \end{subfigure}

    \caption{The illustrative example justifying the "flipping rank" phenomenon. There are three agents with different abilities: Agent 2 (red) with ability 2, Agent 1 (green) with ability 1, and Agent 0.5 (blue) with ability 0.5. Agents can boost their total abilities by gaining half of their connected agents' abilities. The colored values represent the boosted abilities. In four different scenarios, agents must meet the task requirements to succeed. Each successful task rewards 1, outweighing the connection cost of 0.1. Network structures shown are the most efficient, which maximizes performance minus connection costs.}
    \label{fig_simple_example}
\end{figure}

To elucidate these findings, we designed a simplified theoretical example to capture the core aspects of our environment and propose a couple of potential explanations for this "flipping rank" phenomenon, as illustrated in Figure \ref{fig_simple_example}. In this example, agents with varying capabilities, represented by different vision ranges in our experiment, assist one another via connections in a communication network. However, the assistance an agent can provide is typically less than its own capabilities, because the information beneficial to one agent may not be as useful to another. The task difficulty in each episode varies, resulting from the random initial positions of agents and landmarks in our experiment. The system earns rewards when more agents successfully complete tasks, which occurs when their innate abilities combined with the help received from others through the network meet the task's requirements. This scenario represents the combination of agents' own observations and others' information from communication to facilitate solving the landmark spreading task.

In the example, we assume that the cost of establishing one link in the network (0.1) is less than the reward for an additional agent completing the task (1). To resolve a tie when two agents can help another complete the task, the agent with higher ability will offer the help. This decision ensures that our insights remain consistent and are not influenced by random methods such as flipping coins. Here, we have three agents with abilities quantified as 2, 1, and 0.5, which we refer to as agent 2, agent 1, and agent 0.5 respectively. Each agent contributes half of their ability to assist others through undirected connections, facilitating mutual support among connected agents. We then explore the most efficient communication network structures from the system manager's perspective under various scenarios within this simplified framework.

In scenarios where task requirements are minimal (0.5 ability), there is no need for additional links in the communication network since all agents can independently meet these requirements, and adding links would only incur unnecessary costs. For tasks requiring 1 ability, only agent 0.5 cannot solve the task alone. In this case, it is advantageous for agent 2 to connect to agent 0.5. This connection boosts the latter’s total ability to 1.5, enabling it to accomplish the task. Although agent 2 also receives an additional 0.25 ability from agent 0.5, this surplus does not provide further benefit as it already exceeds the task requirement.

When tasks demand 1.5 ability, both agents with abilities of 1 and 0.5 fall short on their own. Thus, agent 2 must establish connections with both to facilitate task completion. In the most challenging scenarios, where the requirement is 2 ability, agent 2 just manages to meet this threshold independently. However, agents with abilities of 1 and 0.5 are unable to complete the task on their own. In this case, agent 0.5 remains unable to fulfill the task requirements despite assistance from agent 2. Consequently, the most effective network structure under these conditions is a fully connected network, where all agents are interconnected, maximizing the distribution of available abilities and support.

Assuming that the tasks with varying difficulties are uniformly distributed across all episodes, the average node degrees for agent 2, agent 1, and agent 0.5 are 1.25, 0.75, and 1, respectively; the average betweenness centralities for agent 2, agent 1, and agent 0.5 are 0.67, 0.29, and 0.54, respectively. The rankings in this illustrative example reveal a similar pattern to our simulation results, showing that the rank of agents' degrees and betweenness centrality may not align with the rank of their abilities. This pattern holds even when the tasks are normally distributed in all episodes, with a higher probability for tasks of middle difficulties, which more closely reflects the conditions of our original experiment.

First, we clearly observe that agent 1 is in a dual role: it needs assistance from others to solve tasks in scenarios requiring abilities of 1.5 and 2, and it can also provide useful help in scenarios requiring an ability of 2. While this factor does not directly result in the "flipping rank" phenomenon, it can become more significant and contribute to this phenomenon in our more complex experiments. Second, the scenario with a requirement of 1 ability is the main reason for this phenomenon. Agent 1 is not powerful enough to assist others significantly, but it can independently solve the task, suggesting it should be isolated from the communication network in some cases. In the scenario with a requirement of 1 ability, even if we relax the initial assumption for breaking ties and instead determine network structures by flipping a coin, the results would not change. 

It should be emphasized that this example significantly simplifies our original environment. In our experiment, quantifying the abilities of agents and the difficulty of tasks is challenging; the process is much more complex, and the assistance offered by agents is heterogeneous, varying with different recipients and evolving over time. However, this example captures the essential aspects and replicates similar patterns observed in our experimental results, potentially explaining the insights behind the "flipping rank" phenomenon as well.

\begin{figure*}
    \centering
    \begin{subfigure}[b]{0.8\textwidth}
        \includegraphics[width=\textwidth]{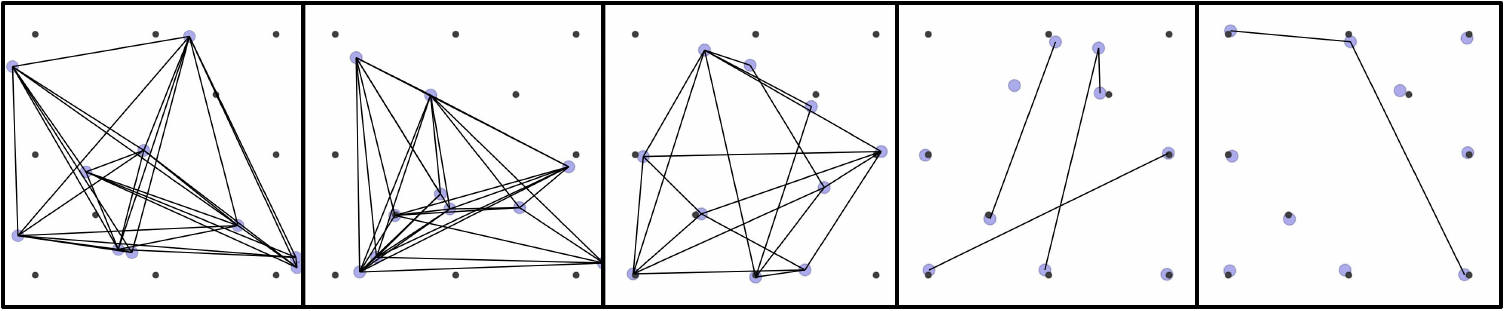}
        \caption{Vision range 0.6}
        \label{fig:first-graph}
    \end{subfigure}
    
    
    \begin{subfigure}[b]{0.8\textwidth}
        \includegraphics[width=\textwidth]{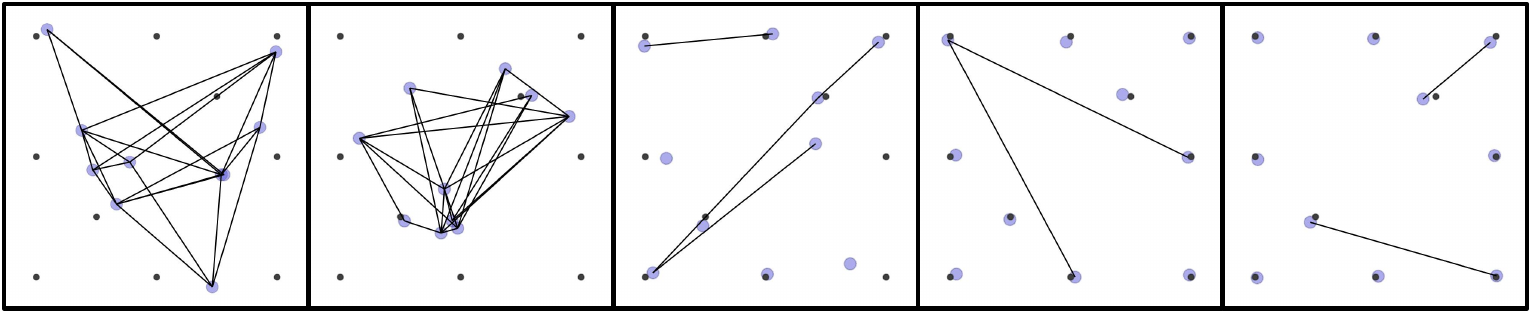}
        \caption{Vision range 0.8}
        \label{fig:second-graph}
    \end{subfigure}
    
    
    \begin{subfigure}[b]{0.8\textwidth}
        \includegraphics[width=\textwidth]{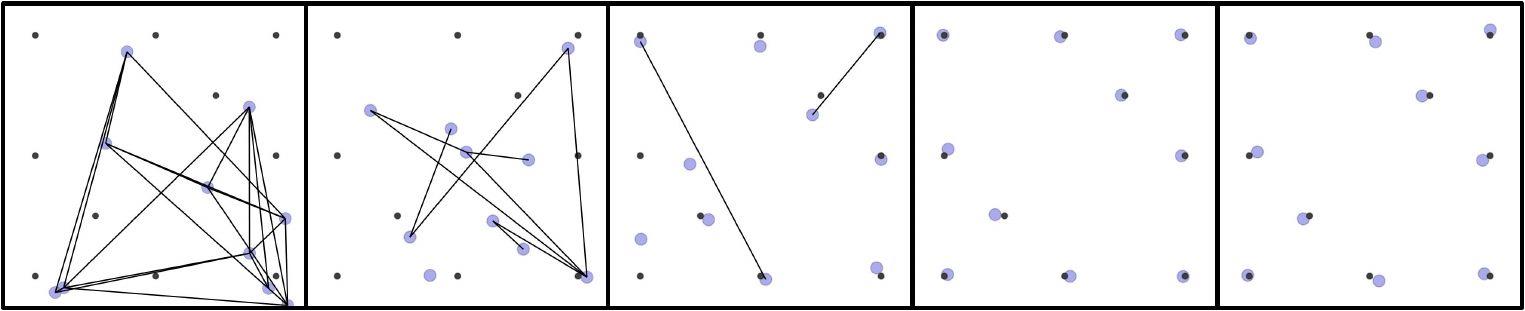}
        \caption{Vision range 1.0}
        \label{fig:third-graph}
    \end{subfigure}
    
    
    \begin{subfigure}[b]{0.8\textwidth}
        \includegraphics[width=\textwidth]{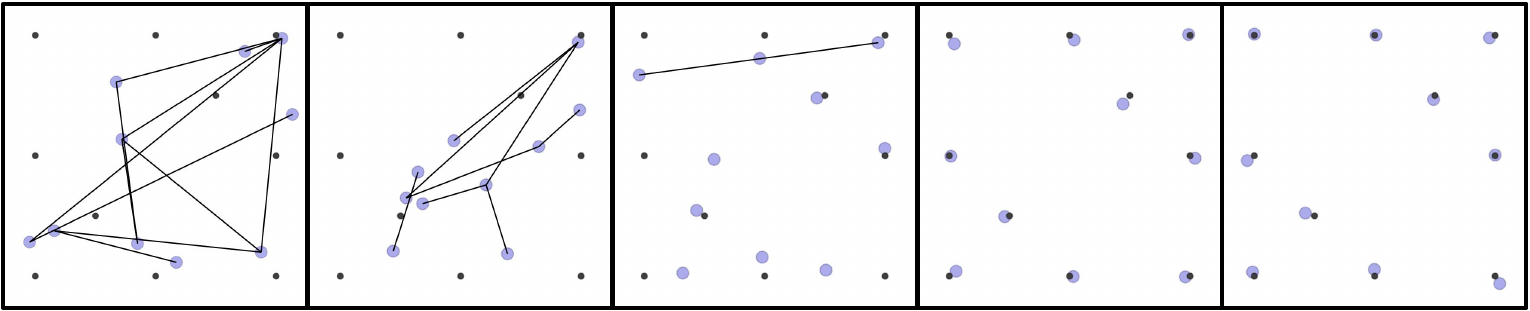}
        \caption{Vision range 1.2}
        \label{fig:fourth-graph}
    \end{subfigure}
    
    \caption{The evolution of communication networks and agents' behaviors using the VAE-RL policy under different vision ranges. The graphs, from left to right, show the system's status at time steps 0, 5, 10, 15, and 20. Each row, from top to bottom, represents the evolution under vision ranges of 0.6, 0.8, 1.0, and 1.2. For clarity, the initial positions of landmarks are set in a grid-like formation, while in other cases, they are randomly initialized.}
    \label{fig:snapshot}
\end{figure*}

\subsubsection{Evolution of network structure}
This visualization aims to elucidate the interactions between agents and the flow of information within communication networks, which is shown in Figure \ref{fig:snapshot}. For this demonstration, we standardize the positions of landmarks while initializing the agents' positions randomly, each with vision ranges of 0.6, 0.8, 1.0, and 1.2. We capture snapshots of the environment at various time steps—0, 5, 10, 15, 20—to observe the progression of the environment and the behaviors of the agents. It should be noted that the training and testing procedures described in our paper incorporate significant randomness to ensure the robustness of our VAE-RL framework. The statistical analysis provided in the previous section offers a more precise evaluation of VAE-RL. The examples presented in this section are intended to visually demonstrate the system's evolution.

We begin by analyzing the evolution of the system over time. Initially, the agents are randomly dispersed across the environment for all scenarios at varying distances from the landmarks. At the early stage, the communication network is notably dense, reflecting our earlier observation that denser networks facilitate information sharing among agents, aiding them in coordinating their efforts and commencing their tasks. From time steps 5 to 15, a gradual thinning of the communication network is observed, with agents progressively moving closer to their landmarks. This trend corroborates our previous finding that communication networks become sparser over time, effectively balancing the dual objectives of task completion and conservation of communication resources. From time steps 15 to 20, the communication network is almost empty for all scenarios, indicating that most agents have either reached or are nearing their landmarks, with only a few agents continuing their journey towards the remaining landmarks. Finally, by time step 20, the snapshot reveals all agents successfully occupying their landmarks, rendering the communication network unnecessary.

We also examined the system's evolution across scenarios with varying vision ranges. A clear trend emerges: as the vision range increases, making the task progressively easier, the communication network among agents becomes sparser at every time step. This trend corroborates our previous statistical analysis findings. Agents with wider vision ranges can more effectively gather information from their surroundings, enhancing their ability to coordinate with others and accomplish tasks independently. Consequently, the reliance on communication decreases since rich communication becomes unnecessary and incurs additional resource costs. For example, the system with a 0.6 vision range maintains a relatively dense communication network up to time step 10, whereas systems with 1.0 and 1.2 vision ranges exhibit sparse communication networks from the outset and completely disable communication after time step 10.

\section{Conclusion}

We introduce VAE-RL, a novel approach for optimizing multi-agent communication networks. Traditional Deep Reinforcement Learning (DRL) struggles with the exponentially large action space of network topologies. VAE-RL addresses this by using a Variational Autoencoder (VAE) to transform the discrete action space into a manageable continuous latent space. DRL algorithms for continuous action spaces (DDPG) then learn policies in this latent space, which are decoded back into network topologies. Tested on a modified OpenAI particle environment \cite{chen2022dynamic}, VAE-RL outperforms baseline methods like Flat-RL and BDQN in both small (4 agents) and large (10 agents) systems, across homogeneous and heterogeneous scenarios. Our analysis reveals insightful trends in VAE-RL's behavior.

While our environment is primarily modeled after the communication networks of multi-robotic systems in the real world, it can also find parallels in other real-world situations. For instance, consider a corporate management system where a department comprises multiple employees, including at least one manager. These employees collaborate to address their respective tasks. However, variations in talent and work experience among the employees lead to differences in task completion performance. In such scenarios, the manager is tasked with making decisions on how to establish connections among employees to facilitate collective task completion, especially for those employees with comparatively weaker abilities. Optimizing departmental performance by strategically assigning connection networks among employees is a critical responsibility of department managers. This scenario closely mirrors the communication network assignment strategy learned by VAE-RL. Notably, our method operates autonomously and is adaptable to systems featuring heterogeneous workers, making it versatile and applicable across various contexts.


The VAE-RL framework holds significant potential for application in numerous real-world multi-agent systems, particularly those reliant on network-based information sharing and interaction processes. However, it has its limitations. A fundamental assumption underpinning VAE-RL is the strong authority of the manager, responsible for assigning network structures to multi-agent systems. This framework presupposes that agents will comply with the manager's commands, a scenario that may not always hold true in real-life situations. Multi-agent systems can be broadly classified into three categories: human-only systems, human-AI systems, and AI-only systems. In the last two categories, the assumption of compliance is more likely to be valid, given the predominance of AI agents programmed to adhere to managerial decisions, except in cases of technical failures. However, in systems with a higher proportion of human agents, this assumption may not always be correct. Even in scenarios where the manager represents high-authority entities such as governments, military, or corporations, human agents may sometimes exhibit rebellious behavior, as evidenced by instances of non-compliance with social distancing and masking policies during the Covid-19 pandemic \cite{he2021people, van2022emergence}. Therefore, while our framework is highly effective in AI-dominated systems, its applicability may be limited in scenarios involving a majority of human agents due to the potential for reduced managerial authority.

Additionally, the proposed VAE-RL framework exhibits a scalability limitation. While we have demonstrated that the framework can effectively manage networks up to 10 nodes and potentially adapt to larger network scenarios, it may struggle with the computational demands posed by real-world applications such as transportation, social media, and human interaction networks, which often comprise thousands of nodes. Although our framework significantly reduces the complexity of the action space in networks, it still faces challenges related to computational expenses. Scaling up engineering management techniques for extensive networks remains an unresolved issue.

However, our framework can accommodate larger networks through the concept of introducing a multi-hierarchical structure \cite{chen2024sos}. The general process may involve dividing the network into several sub-networks using clustering algorithms. A first-layer manager then allocates communication resources, treating each sub-network as a node. Subsequent layers of management distribute these resources to agents within the sub-networks. As the network size increases, this hierarchical structure can be expanded further. This approach introduces new challenges, including the efficiency of network clustering algorithms and the development of super networks that integrate these sub-networks as nodes. Despite these difficulties, this method holds promise for scaling our framework to handle multi-agent systems in significantly larger networks.

\bibliographystyle{unsrt}

\bibliography{arxiv.bib} 

\end{document}